\pdfoutput=1

\documentclass[11pt]{article}

\usepackage{emnlp2021}

\usepackage{float}
\usepackage{times}
\usepackage{latexsym}
\usepackage{graphicx}
\usepackage{subcaption}
\usepackage{amsmath}
\usepackage{amsfonts}
\usepackage[ruled,vlined]{algorithm2e}
\usepackage{xcolor}
\usepackage{enumitem}
\usepackage{multirow}

\usepackage[T1]{fontenc}

\usepackage[utf8]{inputenc}

\usepackage{microtype}

\title{Task-Specific Embeddings for Ante-Hoc Explainable Text Classification}

\author{Kishaloy Halder$^{\dagger}\thanks{~~Work done prior to joining Amazon}$ \hspace{2cm} Josip Krapac \hspace{2cm} Alan Akbik\\
{\bf Anthony Brew}$^{\dagger}$ \hspace{2cm} {\bf Matti Lyra} \\\\
  $^\dagger$ Work done while at Zalando \\\\
  {\small \texttt {kishaloh@amazon.com \hspace{0.2cm} josip.krapac@zalando.de \hspace{0.2cm} alan.akbik@hu-berlin.de}}\\
  {\small \texttt {atbrew@gmail.com \hspace{0.2cm} matti.lyra@zalando.de}} \\
}

\begin{document}
\maketitle
\begin{abstract}

Current state-of-the-art approaches to text classification typically leverage 
BERT-style Transformer models with a softmax classifier, jointly fine-tuned to 
predict class labels of a target task. In this paper, we instead propose an alternative 
training objective in which we learn \textit{task-specific embeddings} of text: our proposed objective
 learns embeddings such that all texts that share the same target class label 
should be close together in the embedding space, while all others should be far apart. 
This allows us to replace the softmax classifier with a more interpretable $k$-nearest-neighbor classification approach. In a series of experiments, we show that this yields a number of interesting benefits: (1) The resulting order induced by distances in the embedding space can be used to directly explain classification decisions.
(2) This facilitates qualitative inspection of the training data, helping us to better understand the problem space and identify labelling quality issues. (3) The learned distances to some degree generalize to unseen classes, allowing us to incrementally add new classes without retraining the model. We present extensive experiments which show that the benefits of ante-hoc explainability and incremental learning come at no cost in overall classification accuracy, thus pointing to practical applicability of our proposed approach. 

\end{abstract}

\section{Introduction}
\label{section:intro}
Text classification is the classic NLP problem of predicting the appropriate class labels for a given textual document from a pre-defined set of classes. It is used for various applications such as sentiment analysis~\cite{rosenthal2017semeval}, spam detection~\cite{jindal2007review} or automatic document categorization~\cite{zhang2015character}. The current state-of-the-art approach leverages BERT-style language models together with a softmax classifier~\cite{devlin2019bert,wang2019glue}. The language model is fine-tuned using a task's training data to produce a vector representation of a given text, typically retrieved from the \textsc{[cls]} token of the language model. This representation is trained such that a simultaneously trained softmax classifier projects it into a distribution over class label probabilities. 

In this work, we explore a complementary path to address the text classification 
problem which we argue, yields significant advantages in terms of explainability and incremental learning, while giving similar results in terms of classification accuracy. 

\begin{figure*}[t]
    \vspace{-1mm}
    \centering
    \begin{subfigure}[t]{0.5\textwidth}
        \centering
        \includegraphics[height=1.5in]{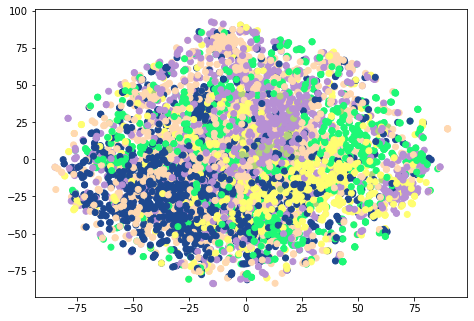}
        \caption{Sentence-BERT}
        \label{subfig:stsb}
    \end{subfigure}%
    ~
    \begin{subfigure}[t]{0.5\textwidth}
        \centering
        \includegraphics[height=1.5in]{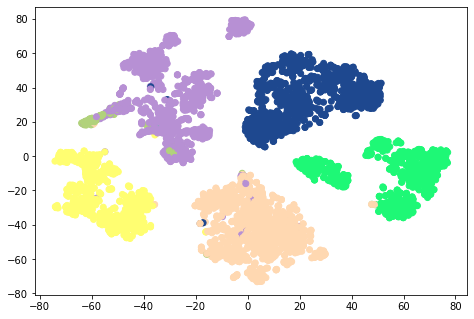}
        \caption{CEA (our proposed)}
        \label{subfig:cea}
    \end{subfigure}
    \caption{TSNE plot of embeddings for training samples in \textsc{TREC-6} corpus, obtained from a) Sentence-BERT: A generic Semantic Textual Similarity model with strong performance on NLI and STS tasks, and b) Our proposed \textbf{C}lass-driven \textbf{E}mbedding \textbf{A}lginment (CEA) model. Colors indicate the class of a point. The embedding space is visibly aligned well with the downstream class labels in case of CEA, though subcluster structures are clearly visible, as well as pockets of differently colored points.}
    \label{fig:tsne}
    \vspace{-2mm}
\end{figure*}

\noindent
\textbf{Task-specific similarity of texts.}
We propose to view text classification as a task-specific text-text similarity problem. In other words, we assume that if two textual documents share the same class label, then they are semantically similar in the context of this particular task. To illustrate, consider the following two sentences from a question type classification task: 
\begin{itemize}
    \item \textit{Which one of the Great Lakes is entirely within U.S. territory?}
    \item \textit{What arch can you see from the Place de la Concorde?}
\end{itemize}
These two sentences seemingly have little in common in terms of semantics\footnote{In fact, when embedded with a generic STS model {\tt stsb-roberta-base} (\citeauthor{reimers2019sentence}) we find only a cosine similarity of $0.28$ between their embeddings.}.
However, they both share the same label in a specific downstream task (namely {\tt LOC}, indicating a question that requires a location name as answer) and so could be argued to be similar when considering only the semantics of the target task.

Based on this observation, we propose a training objective called \textbf{C}lass-driven \textbf{E}mbedding \textbf{A}lignment (CEA). The main idea is to learn embeddings that maximize the similarity between two textual documents if they share the same class label, and minimize it if they do not. This builds on recent work in learning representations for semantic textual similarity (STS) using Siamese networks~\cite{reimers2019sentence}, but with the difference that we aim to learn task-specific embeddings instead of broad semantic representations. We illustrate the difference between the two paradigms in Figure~\ref{fig:tsne} showing a TSNE plot of an embedding space learned for general semantics (\ref{subfig:stsb}) vs. one learned for task-specific representations of semantics  (\ref{subfig:cea}).

\noindent
\textbf{Advantages for text classification.}
We show that such a task-specific embedding space can directly be used for text classification and that this yields a number of desirable properties: 

\begin{enumerate}[leftmargin=*]
    \item We show that task-specific embeddings can be employed in a $k$-nearest-neighbor (kNN) approach to classify texts based on their similarity to training data points. As kNN is an instance of example-based learning, this allows \textit{ante-hoc explainations} of the classification results: each classification decision is based on the identified nearest neighbors from the training data and thus is explained in a human-readable way. 
    \item This facilitates qualitative exploration of training data to better understand the problem space and to identify labeling quality issues in the training data. For instance, our example TSNE plot in Figure~\ref{subfig:cea} reveals that task-specific embeddings mostly form clusters conforming to their class labels, but not entirely (note the subclusters and the pockets of differently colored data points). As we show in Section~\ref{section:discussion}, exploration reveals pockets of wrongly labeled data points in the training data of some tasks, as well as other anomalies.
    \item Finally, the proposed formulation is also effective for incremental learning approaches in which more labeled data points are added after training. We show that the similarities in the learned embedding space partially generalize even to new class labels, allowing us to add new classes without retraining.
\end{enumerate}

We present extensive quantitative and qualitative experiments over $6$ datasets from various text classification tasks that indicate our formulation yields these properties with no hit in overall classification accuracy. Based on these results, we conclude our proposed approach to be a viable alternative with many benefits over standard softmax classifier-based approaches.

\section{Method}
\vspace{-1mm}
\label{section:method}
We formulate the text classification problem as an application-specific text-text
similarity problem. In the following, we would use the terms sentence and document 
interchangeably to refer to a particular text input to be classified. Formally, 
we aim to learn a function:

\begin{equation}
\label{equation:metric_learning}
\begin{gathered}
    f\colon text \to \mathbb{R}^{d} \\
    f(text) \approx f(text^{+}) ; \quad f(text) \ne f(text^{-}) 
\end{gathered}
\end{equation}

where, $text^{+}$ is another document which shares the same class
as that of $text$, and $text^{-}$ is one which does not; 
$\approx$ (or $\ne$) denote equality (or inequality) in terms of some similarity metric. 
In other words, we aim to learn a text encoder which embeds a textual 
document in a $d$-dimensional vector space. In this space, documents 
that share the same class should be close to each other, and the ones
from different classes should be 
further apart.

\subsection{Sampling for CEA Training}
\label{subsec:cea-sampling}
To train task-specific embeddings with CEA, we require input pairs (\textit{e.g.,} $\langle text, text^{+} \rangle$, $\langle text, text^{-}\rangle$) 
as shown in Equation \ref{equation:metric_learning}. Since with $n$ labelled samples in the corpus the number of pairs grows quadratically, we sample pairs from this set of $\frac{n(n-1)}{2}$ possible combinations over epochs, where each epoch consists of $n$ steps as shown in Algorithm \ref{algo:cea-sampling}. 
Specifically, for every $text$ document, we pick a positive sample document $text^{+}$ 
with equal probability from all other documents in the training set that share the 
same class as of $text$ document. Similarly, we pick a negative sample $text^{-}$ 
from the pool of all documents that belong to different classes than that of 
$text$ document. We repeat this sampling process for all the documents in an 
epoch, resulting in $2n$ document pairs. Note that as we sample the pairs 
randomly, in each epoch different positive, and negative examples might be 
picked for a particular document.

\begin{algorithm}[t]
\small
\SetAlgoLined
\KwIn{Labelled documents $\{(text_i, label_i) \forall i \in [1, n]\}$}
\KwOut{Positive and negative sample pairs}
 $samples \gets \{ \}$\;
\For{$i \gets 1$ to $n$}{
    $text \gets text_i$\;
    $text^{+} \gets \sim~\mathcal{U}(\{text_j\forall j\mid label_j=label_i\})$\;
    $text^{-} \gets \sim~\mathcal{U}(\{text_j\forall j\mid label_j\ne label_i\})$\;
    
    $samples \gets samples \uplus \langle(text, text^{+}), 1\rangle$\;
    $samples \gets samples \uplus \langle(text, text^{-}), 0\rangle$\;
}
 return $samples$
 \caption{Sampling for CEA training.}
 \label{algo:cea-sampling}
\end{algorithm}

\subsection{CEA Training}
\label{subsec:cea-training}
\begin{figure}[t]
    \centering
    \includegraphics[height=1.8in]{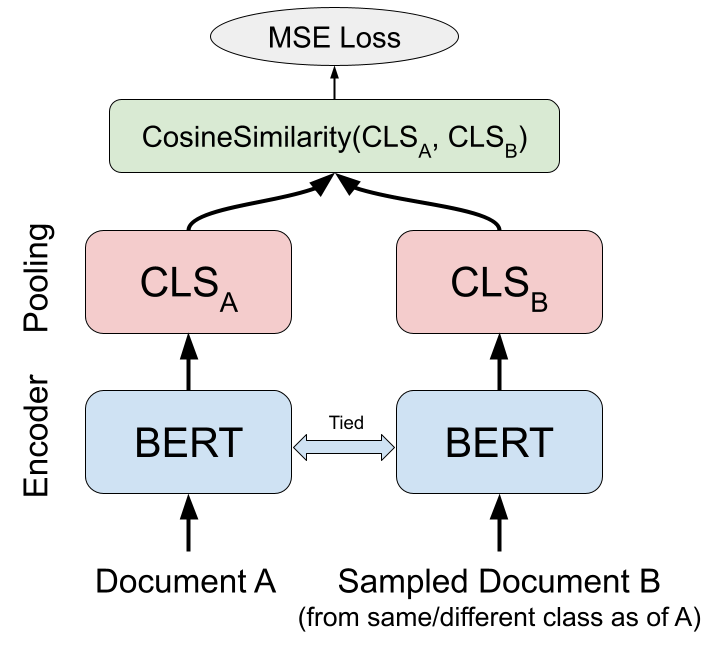}
    \caption{Siamese architecture for Class-driven Embedding Alignment (CEA) training.}
    \label{fig:cea_arch}
    \vspace{-5mm}
\end{figure}

We use a Siamese architecture for CEA training as shown in Figure \ref{fig:cea_arch}.
It uses a pre-trained BERT as the text encoder, and the weights for both the BERT towers
are tied. We consider every positive sample pair $\langle text, text^{+} \rangle$ to 
have a target label of $1$, conversely negative sample pair 
$\langle text, text^{-} \rangle$ has target label of $0$. The network 
takes a sample pair as input, makes a forward pass through the BERT stack for both
the input documents, and yields the sub-token level encoded representations from 
the final layer of the stack.

{\noindent \textbf{Deriving a document representation.}} There exist several strategies to derive document-level
representations from BERT-style language models, such as using the representation at the \textsc{[cls]} token or performing mean or max pooling over all subtoken representations.  \citeauthor{reimers2019sentence} showed the effect of different pooling mechanisms and found mean pooling to yield best results for their tasks. However, as we are primarily interested in performing text classification over documents with potentially widely varying lengths, we follow current practice in using the \textsc{[cls]} token representation of the final output layer as document embedding. 

{\noindent \textbf{Loss calculation.}} For each embedded input pair, we compute the \textit{MeanSquaredError} (MSE) between their \textit{CosineSimilarity} 
value $\hat{y}$ and the target value $y$: 

\begin{equation}
\begin{gathered}
    \hat{y} = CosineSimilarity( \text{CLS}_\text{A}, \text{CLS}_\text{B} ) \\
    loss = MeanSquaredError(y, \hat{y})
\end{gathered}
\end{equation}

The target value  $y$ is either $1$ or $0$ depending on if both the inputs share the same class or not.
This objective thus pushes documents with the same labels to be close together in the embedding space (high cosine similarity) and all others far apart.

\subsection{Nearest Neighbor Classification}
\label{subsec:cea-nn}
After training CEA, we perform classification for unseen documents using a $k$-nearest neighbor approach. This requires us to first encode all documents in the training data with the task-specific CEA model, yielding $n$ $d$-dimensional embeddings. We normalize these embeddings with the respective $L2$ norms to obtain unit-norms, and store all embeddings in a look-up index.

For inference, we encode a given data point with the CEA model and normalize the resulting embedding to have unit-norm. We fetch $k$ nearest points in the look-up index in terms of $L2$ distance, and consider the majority class label as the predicted class for the document under test.

{\noindent \textbf{Inference speed.}} The inference time of $k$NN is
theoretically linear with respect to the size of the lookup index. In practice, 
it is possible to accelerate the look-up using efficient algorithms
such as \textit{Ball Tree}~\cite{omohundro1989five}, parallelization and approximate NN methods  \cite{JDH17}.

\section{Evaluation of Classification Accuracy}
\label{section:experiments}

In this first round of evaluation, we investigate whether (1) our proposed approach combining CEA, and $k$NN classification comes at a cost of accuracy compared to state-of-the-art approaches, and (2) whether CEA can additionally be used in combination with traditional softmax-classifier instead of $k$NN. In Section~\ref{section:discussion}, we examine CEA further, regarding the postulated advantages \textit{i.e.,} better intepretability and incremental learning.

\subsection{Datasets}
\label{subsec:datasets}

We consider six widely used datasets in English from three domains, formally released with \textit{training} and \textit{test} splits (\textit{cf} Table \ref{tab:dataset_stats}). In all cases, we consider a random subset ($10\%$) from the \textit{training} as the development set . We use identical splits for all evaluated models and baselines. The original \textit{test} set remains untouched and is used as-is for evaluation.

\noindent{\textbf{Question type detection.} \textsc{TREC} is a corpus annotated
with coarse and fine-grained question types~\cite{li2002learning}. There are $50$ fine-grained question types (\textsc{TREC-50}) that each belong to one of $6$ coarse-grained categories (\textsc{TREC-6}). The two layers of annotations allow us to investigate whether CEA uncovers structure beyond the class labels it is trained with.}

\noindent{\textbf{Sentiment analysis.} Two popular review datasets, on movies (\textsc{IMDb}), and restaurants (\textsc{Yelp-full}). They are annotated with coarse (binary positive/negative), and fine-grained $5$-class sentiment labels~\cite{zhang2015character, maas2011learning} respectively. Documents are longer in general here.}

\noindent
\textbf{Topic detection.} Two commonly used topic classification datasets, namely Wikipedia-based (\textsc{DBPedia}) and news articles (\textsc{AGNews}), with $14$ and $4$ topic labels respectively \cite{zhang2015character}.

\subsection{Proposed Model and Ablations}
\label{subsec:baselines}

For each dataset, we start with a pre-trained $\text{BERT}_\text{base}$ model ($110M$ parameters), and perform our proposed CEA training as described in Section \ref{section:method} using {\tt AdamW} optimization \cite{loshchilov2018decoupled}, a batch size of $16$\footnote{the siamese architecture has to fit double amount of data in memory for each forward pass by design, hence smaller batch-size.} and a learning rate of 2$e$-5 for $20$ epochs. We follow~\citet{reimers2019sentence} and perform model selection based on \textit{Spearman Correlation} between $\hat{y}$, $y$ on held-out development set. 

\begin{table}[t]
\centering
\addtolength{\tabcolsep}{-3pt}
\resizebox{0.5\textwidth}{!}{%
\begin{tabular}{llccccc}
\hline
Dataset & Type & \#class & \#train / \#test & \#words \\ \hline
 \textsc{TREC-6} \citeyearpar{li2002learning} & Question & 6 & 5.5k / 500 & 11\\  
 \textsc{TREC-50} \citeyearpar{li2002learning} & Question & 50 & 5.5k / 500 & 11\\  
 \textsc{IMDb} \citeyearpar{maas2011learning}& Sentiment & 2 & 25k / 25k & 190 \\
 \textsc{Yelp-full} \citeyearpar{zhang2015character} & Sentiment & 5 & 650k / 50k & 136 \\ 
 \textsc{AGnews} \citeyearpar{zhang2015character} & Topic & 4 & 120k / 7.6k & 37\\ 
 \textsc{DBPedia} \citeyearpar{zhang2015character} & Topic & 14 & 560k / 70k & 49\\ \hline
\end{tabular}%
}
\caption{Dataset statistics.}
\label{tab:dataset_stats}
\vspace{-5mm}
\end{table}

\noindent
\textbf{CEA-based classification.} We show the use of CEA model in three following methods:

\begin{enumerate}[leftmargin=*]

    \item {\textbf{CEA + $k$NN}: Our proposed model uses $k$NN classification as described in Section \ref{subsec:cea-nn}. To find the optimal $k$ for each dataset, we perform a 
     hyper-parameter search for $k$ within the range $[1-100]$, and select the best $k$ using accuracy on the development split. Unseen documents are classified with majority voting over the $k$ nearest neighbors.}
    
    \item {\textbf{CEA + Softmax}:} An ablation where instead of $k$NN classification, we add a softmax-classifier to the trained CEA model and fine-tune all model parameters on the target task using cross-entropy loss, as typically done in text classification. The purpose is to evaluate whether CEA provides a good initialization of model parameters for the standard text classification approach. 
    
    \item {\textbf{CEA (frozen) + Softmax}: Another ablation that fits a softmax-classifier, but freezes all weights in the CEA model. We perform this experiment to determine if it can be directly used to learn a softmax classifier.}
\end{enumerate}

\begin{table*}[t]
\centering
\resizebox{\textwidth}{!}{%
\begin{tabular}{lcccccc}
\hline
Methods & \textsc{TREC-6} & \textsc{TREC-50} & \textsc{IMDB} & \textsc{DBPedia} & \textsc{AGNews} & \textsc{Yelp-full}\\
\hline
1. CEA + $k$NN & $\mathbf{0.9706} {\scriptstyle \pm .0049}$ & $0.8980 {\scriptstyle \pm .0028}$ & $0.9314 {\scriptstyle \pm .0005}$ & $\mathbf{0.9932} {\scriptstyle \pm .0000}$ & $0.9366 {\scriptstyle \pm .0013}$ & $0.6916 {\scriptstyle \pm .0009}$\\
2. CEA + Softmax & $0.9680 {\scriptstyle \pm .0028}$ & $\mathbf{0.9093} {\scriptstyle \pm .0049}$ & $\mathbf{0.9344} {\scriptstyle \pm .0006}$ & $0.9930 {\scriptstyle \pm .0001}$ & $\mathbf{0.9387} {\scriptstyle \pm .0007}$ & $0.6955 {\scriptstyle \pm .0001}$ \\

3. CEA (f) + Softmax & $0.9653 {\scriptstyle \pm .0033}$ & $0.8840 {\scriptstyle \pm .0090}$ & $0.9329 {\scriptstyle \pm .0012}$ & $0.9929 {\scriptstyle \pm .0001}$ & $0.9352 {\scriptstyle \pm .0007}$ & $0.6890 {\scriptstyle \pm .0017}$ \\

\hline \hline
4. S-BERT (f) + Softmax & $0.8286 {\scriptstyle \pm .0002}$ & $0.7120 {\scriptstyle \pm .0090}$ & $0.8719 {\scriptstyle \pm .0004}$ & $0.9788 {\scriptstyle \pm .0004}$ & $0.8891 {\scriptstyle \pm .0006}$ & $0.5831 {\scriptstyle \pm .0005}$ \\
5. BERT + Softmax & $0.9626 {\scriptstyle \pm .0018}$ & $0.9013 {\scriptstyle \pm .0041}$ & $0.9286 {\scriptstyle \pm .0012}$ & $0.9919 {\scriptstyle \pm .0002}$ & $0.9362 {\scriptstyle \pm .0002}$ & $\mathbf{0.6972} {\scriptstyle \pm .0005}$ \\
\hline
\end{tabular}%
}
\caption{Results on text classification datasets. Our CEA variants outperform other baselines with no additional data used during training across the board (except \textsc{Yelp-full}).}
\label{tab:classification_results}
\vspace{-5mm}
\end{table*}

\subsection{Baselines}
\label{subsec:baselines}

We compare against the following baseline approaches based on the same $\text{BERT}_\text{base}$ model

\begin{enumerate}[leftmargin=*]
\setcounter{enumi}{3}
    \item {\textbf{S-BERT (frozen) + Softmax} \cite{reimers2019sentence} is a BERT model trained on STS and NLI to capture generic semantic textual similarity. We use the pre-trained model\footnote{\url{huggingface.co/sentence-transformers/bert-base-nli-stsb-mean-tokens}} 
    released by the authors. As recommended, we use the mean-pooled 
    embedding as the document representation and train a softmax classifier 
    on top in this method. This lets us compare our task-specific representations against general semantic representations.}
    
    \item {\textbf{BERT + Softmax} \cite{devlin2019bert}: In this, we fine-tune  $\text{BERT}_\text{base}$ directly on the task, together with a standard softmax classifier. This experiment benchmarks how well the traditional text classification approach fares with ours.}

\end{enumerate}

All model variants with softmax classifier are fine-tuned for $20$ epochs using batch-size of $32$, and learning rate of 3$e$-6. Model selection is performed on development data accuracy.
We implement the variants using: \textsc{Sentence-Transformer}
for CEA training 
\cite{reimers2019sentence}, \textsc{Flair}
for fine tuning 
\cite{akbik2019flair}, and \textsc{Faiss}
for $k$NN look-up \cite{JDH17}.

\subsection{Model Accuracy Results}
\label{subsec:main_results}

For all models and datasets, we report accuracy on the held-out test set. Since Transformer models are 
known for their high variance \cite{dodge2020fine, halder-etal-2020-task, mosbach2021on} we train $3$ independent models with controlled random seeds, and report the mean and standard deviation. The results are presented in Table \ref{tab:classification_results}.

We observe that our CEA approach performs competitively with other variants. 
In $5$ out of $6$ datasets, CEA based models are able to outperform the Sentence-BERT, and BERT fine-tuned models. Unsurprisingly, the improvement is nominal as the underlying model (thus number of learnable parameters in the Transformer stack), as well as the ground-truth data used for training are exactly the same. Although Sentence-BERT model is pre-trained on supervised tasks such as STS, NLI, and in general is reported in the literature 
to be able to capture semantic similarity well, for downstream text classification tasks, these generic sense of similarity is not effective as evident from the results.

Our proposed CEA approach conditions the embedding space well to show strong classification performance. CEA 
embeddings can be seamlessly used in conjunction with a softmax classifier on top (CEA + Softmax, CEA (f) + Softmax), as well as with a $k$-nearest neighbour lookup (CEA + $k$NN). We observe that we are able to obtain the best results when fine-tune the CEA obtained model further with a softmax classifier on top (CEA + Softmax). This is expected since, it has additional learnable parameters. In case of \textsc{Yelp-full}, the BERT fine tuned model achieves the best results, closely followed by our CEA variants. From these observations, we conclude that CEA indeed provides a viable mechanism to train text classification model with strong performance on a range of tasks.

\begin{table}[]
\centering
\resizebox{0.45\textwidth}{!}{%
\begin{tabular}{lccc}
\hline
Dataset & \begin{tabular}[c]{@{}c@{}}Avg.\\ Runtime (ms)\end{tabular} & \begin{tabular}[c]{@{}c@{}}Avg.\\ \#word\end{tabular} & Size of index\\ \hline
\textsc{TREC-6/50} &  $0.4$ & $11$ & $5.5k$\\
\textsc{AGNews} & $2.9$ & $37$ & $120k$\\
\textsc{IMDb} & $9.5$ & $190$ & $25k$\\
\textsc{DBPedia} & $2.5$ & $49$ & $560k$\\
\textsc{Yelp-full} & $10$ & $136$ & $650k$ \\ \hline
\end{tabular}%
}
\caption{Inference speed of CEA based text classification. The average time taken in a batch inference
setting is reasonably fast.}
\label{tab:inference-speed}
\vspace{-5mm}
\end{table}
\subsection{Inference Speed of CEA}
\label{subsec:inference-speed}
With our proposed CEA + $k$NN approach, two major operations take place for each input document \textit{i.e.,}
(i) computing the {\tt [CLS]} token's embedding from the Transformer stack; (ii) retrieving the $k$ most similar
points in nearest neighbor index to perform majority voting. We present the average running time 
(total time/number of test points) in a batch inference setting on Nvidia V100 GPUs in Table \ref{tab:inference-speed}.
We observe that the inference is reasonably fast with average running time being less than $10$ms for all the datasets.
In general, the running time appears to depend mostly on the average length of the documents rather than the
number of indexed points.

\begin{table*}[ht]
    \begin{subtable}[h]{0.5\textwidth}
        \centering
        \addtolength{\tabcolsep}{-5pt}
        \resizebox{1\textwidth}{!}{%
        \begin{tabular}{lll}
        \hline
        \begin{tabular}[c]{@{}l@{}}Test\\ Document:\end{tabular} & ``What is the sales tax in Minnesota?'' & \begin{tabular}[c]{@{}l@{}}{\color{red} True label: ENTY} \\ {\color{blue} Predicted: NUM}\end{tabular} \\ \hline
        Nearest \#1 & \begin{tabular}[c]{@{}l@{}}``What will the California gas tax be \\ in the year 2000?''\end{tabular} & {\color{blue}True label: NUM} \\ \hline
        Nearest \#2 & \begin{tabular}[c]{@{}l@{}}``What is the fine for having a dog on \\ a beach ?''\end{tabular} & {\color{blue}True label: NUM} \\ \hline
        Nearest \#3 & \begin{tabular}[c]{@{}l@{}}``What are bottle caps with presidents' \\ pictures inside worth?''\end{tabular} & {\color{blue}True label: NUM} \\ \hline
        Nearest \#4 & \begin{tabular}[c]{@{}l@{}}``What is the per-capita income of \\ Colombia, South America?''\end{tabular} & {\color{blue}True label: NUM} \\ \hline
        \end{tabular}%
        }
        \caption{Labelling inconsistency.}
        \label{tab:label-inconsistency}
    \end{subtable}
    ~
    \begin{subtable}[h]{0.5\textwidth}
        \centering
        \addtolength{\tabcolsep}{-5pt}
        \resizebox{1.07\textwidth}{!}{%
        \begin{tabular}{lll}
        \hline
        \begin{tabular}[c]{@{}l@{}}Test\\ Document:\end{tabular} & \begin{tabular}[c]{@{}l@{}} ``What is the electrical output in Madrid,\\ Spain?''\end{tabular} & \begin{tabular}[c]{@{}l@{}}{\color{red}True label: ENTY} \\ {\color{blue}Predicted: NUM}\end{tabular} \\ \hline
        Nearest \#1 & \begin{tabular}[c]{@{}l@{}}``What are the unemployment statistics \\ for the years 1965 and 1990?''\end{tabular} & {\color{blue}True label: NUM} \\ \hline
        Nearest \#2 & \begin{tabular}[c]{@{}l@{}}``What are the unemployment statistics \\ for the years 1965 and 1990?''\end{tabular} & {\color{blue}True label: NUM} \\ \hline
        Nearest \#3 & \begin{tabular}[c]{@{}l@{}}``What is the probability that at least 2 out \\ of 25 people will have the same birthday?''\end{tabular} & {\color{blue}True label: NUM} \\ \hline
        Nearest \#4 & \begin{tabular}[c]{@{}l@{}}``What is the probability that at least 2 out \\ of 25 people will have the same birthday?''\end{tabular} & {\color{blue}True label: NUM} \\ \hline
        \end{tabular}%
        }
        \caption{Duplicates.}
        \label{tab:duplicate-documents}
    \end{subtable}
    \caption{Case studies of data points that were mis-classified by our CEA based $k$-nearest neighbor method. The nearest neighbors might be considered justification for the prediction outcome. Those hint at: (a) potential labelling inconsistencies, (b) Duplicates in the formally released dataset.}
    \label{table:case-study}
    \vspace{-5mm}
\end{table*}
\section{Evaluation of Interpretability and Incrementality}
\label{section:discussion}

In this section, we answer important claims regarding the useful properties of CEA embeddings, 
we made in Section \ref{section:intro}.

\subsection{Transparency in Classification Process}
\label{subsec:transparency}
Our training objective of bringing documents with the same class label 
closer in the embedding space, helps the $k$NN classifier to rely on 
the neighbor sentences to achieve strong performance. We perform a qualitative 
study with the retrieved nearest neighbors, and obtain some interesting 
insights. A few case studies where our CEA + $k$NN method's predictions 
differ from the ground-truth labels, are presented in Table \ref{table:case-study} 
for \textsc{TREC-6}.

First, we observe that the retrieved neighbor sentences indeed share similarity
in the context of the downstream task \textit{i.e.,} question type detection. We believe 
these nearest points can be considered as a form of justification of the model 
prediction. As shown in Table \ref{tab:label-inconsistency}, the test document 
\textit{``What is the sales tax in Minnesota?''} is classified as ``NUM''
by our model. Looking at the supporting justifications retrieved, it is hard to 
argue why this classification is incorrect. Second, this can also help in making 
corrections in the labelled corpus, as one can just take a note of such cases, and 
make corrections in the corpus. Another such example is provided in Table \ref{tab:duplicate-documents}.
Here, not only the ground-truth label ``ENTY'' for the document \textit{``What is the 
electrical output in Madrid, Spain?''} is questionable, but the retrieved nearest 
neighbors highlight that there are duplicates in this \textit{formally released} 
dataset. Both of these observations depict an interesting, and useful artifact of 
our model, that it points at such anomalies in a labelled corpus by design, which 
could be a time-consuming exercise to spot otherwise.

\subsection{Incremental Learning Capabilities}
\label{subsec:incremetal-learning}
Our proposed method uses $k$-nearest neighbor search which is a \textit{lazy classification} 
approach. It decouples the final prediction from the training step, and allows one
to just \textit{add} more knowledge in form of labelled examples in the search index. This
introduces a significant advantage for human-in-the-loop oriented systems, where newly 
labelled data arrives periodically. Traditionally, it would require a full-fledged model
training loop to utilize the additional labelled data, which might be expensive in terms of
time and compute resources. In the following two experiments, we show that once our
CEA model is trained on part of the corpus, it would allow one to incrementally add more 
labelled data into the index, and still get comparable performance with respect to a model
trained from scratch on all the data points.\\

\begin{figure}[t]
    \centering
    \includegraphics[height=2in]{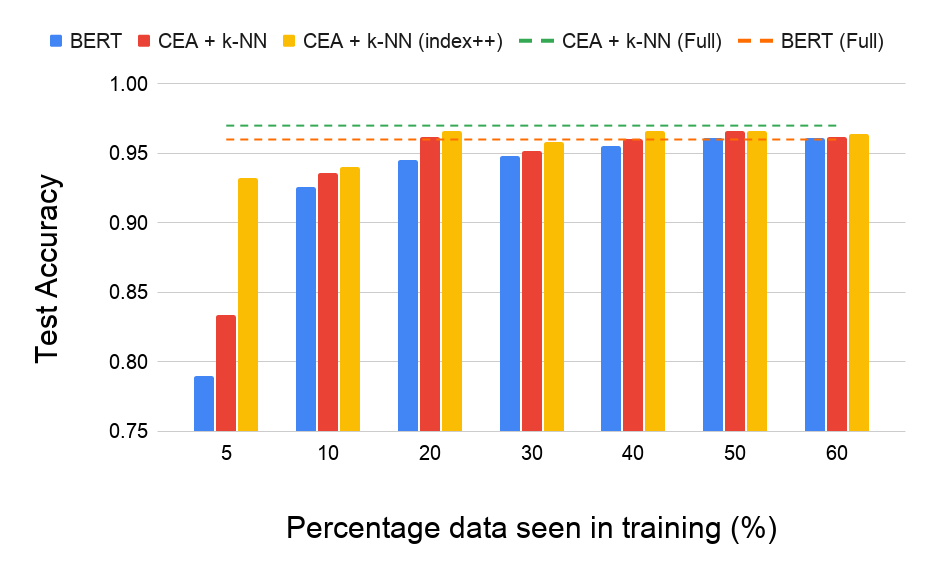}
    \caption{Incremental Learning capabilities of CEA when more labelled data is 
    made available to an already trained model.}
    \label{fig:incremetal-learning-more-data}
    \vspace{-5mm}
\end{figure}

\noindent{\textbf{Introducing more data incrementally.}}
To simulate the incremental data arrival scenario, we train CEA models with $x\%$ of all the
labelled documents available in the training corpus. We vary $x$ between $[5, 10, 20, ..., 60]$
and note the accuracy on the entire test set for \textsc{TREC-6} corpus, at all stages as 
shown in Figure \ref{fig:incremetal-learning-more-data}. We ensure that all the classes appear
at least once in training for all $x$. It is clear from Table \ref{tab:classification_results} 
that the Sentence-BERT is not effective in this setting. Therefore we do not display it in 
Figure \ref{fig:incremetal-learning-more-data} since with $60\%$ of training data it reaches accuracy of only $0.75$. Once we add the rest $(100-x)\%$ labelled documents to the index, denoted by CEA + $k$NN (index++), we observe a boost in the performance compared to both CEA + $k$-NN, and fine-tuned BERT variant. The accuracy of CEA + $k$NN at small values of $x$ is quite close (within $3-4\%$) to what a model would achieve if trained on $100\%$ of the available data, denoted by CEA + $k$-NN (Full), and BERT (Full). Unsurprisingly, the improvement subsides as we move beyond $x\ge50\%$.

\begin{table*}[t]
\centering
\addtolength{\tabcolsep}{-3pt}
\begin{tabular}[t!]{cc}
\begin{minipage}[t!]{.5\textwidth}
\resizebox{0.9\textwidth}{!}{%
\begin{tabular}{p{2.8cm}clcc}
\multicolumn{5}{c}{Corpus: \textsc{TREC-6} \hfill Held-out Class: ``ABBR''} \\ \hline
 & Held-out & Index++ & Full & Support \\ \hline
ABBR & $-$ & $0.7500\uparrow$ & $0.8750$ & $9$ \\
DESC & $0.9484$ & $0.9718\uparrow$ & $0.9750$ & $138$ \\
ENTY & $0.9101$ & $0.9101$ & $0.9450$ & $94$ \\
HUM & $0.9697$ & $0.9697$ & $0.9767$ & $65$ \\
LOC & $0.9753$ & $0.9753$ & $0.9816$ & $81$ \\
NUM & $0.9912$ & $0.9912$ & $0.9869$ & $113$ \\ \hline
Accuracy & $0.9500$ & $0.9620$ & $0.9720$ & $500$ \\ \hline \\
\end{tabular}%
}%

\resizebox{.9\textwidth}{!}{%
\begin{tabular}{p{2.8cm}clcc}
\multicolumn{5}{c}{Corpus: \textsc{AGNews} \hfill Held-out Class: ``World''} \\ \hline
 & Held-out & Index++ & Full & Support \\ \hline
World & $-$ & $0.9046\uparrow$ & $0.9484$ & $1900$ \\
Sports & $0.9227$ & $0.9744$ & $0.9842$ & $1900$ \\
Business & $0.7555$ & $0.8883\uparrow$ & $0.9041$ & $1900$ \\
Sci/Tech & $0.7602$ & $0.8993\uparrow$ & $0.9057$ & $1900$ \\ \hline
Accuracy & $0.7073$ & $0.9168$ & $0.9356$ & $7600$ \\ \hline \\
\end{tabular}%
}
\vspace{-3mm}
\end{minipage}
&
\begin{minipage}[t!]{.5\textwidth}
\vspace{-15.5pt}
\resizebox{0.9\textwidth}{!}{%
\begin{tabular}{lclcc}
\multicolumn{5}{c}{Corpus: \textsc{DBPedia} \hfill Held-out Class: ``Company''} \\ \hline
& Held-out & Index++ & Full & Support \\ \hline
Company & $-$ & $0.9655\uparrow$ & $0.9776$ & $5000$ \\
Animal & $0.9978$ & $0.9983$ & $0.9989$ & $5000$ \\
Plant & $0.8797$ &  $0.9976\uparrow$ & $0.9980$ & $5000$ \\
Album & $0.9946$ & $0.9961$ & $0.9966$ & $5000$ \\ 
Film & $0.9851$ & $0.9951$ & $0.9961$ & $5000$ \\ 
WrittenWork & $0.9705$ & $0.9921$ & $0.9938$ & $5000$ \\ 
EducationalInst. & $0.9571$ & $0.9880$ & $0.9893$ & $5000$ \\ 
Artist & $0.8945$ & $0.9839\uparrow$ & $0.9895$ & $5000$ \\ 
Athlete & $0.9948$ & $0.9969$ & $0.9975$ & $5000$ \\ 
OfficeHolder & $0.9759$ & $0.9855$ & $0.9900$ & $5000$ \\ 
MeanOfTransport. & $0.9476$ & $0.9943$ & $0.9957$ & $5000$ \\ 
Building & $0.8871$ & $0.9832\uparrow$ & $0.9870$ & $5000$ \\ 
NaturalPlace & $0.9928$ & $0.9970$ & $0.9971$ & $5000$ \\ 
Village & $0.9985$ & $0.9989$ & $0.9986$ & $5000$ \\ 
\hline
Accuracy & $0.9237$ & $0.9909$ & $0.9932$ & $70000$ \\ \hline
\end{tabular}%
}
\vspace{-3mm}
\end{minipage}
\end{tabular}

\caption{Incremental learning abilities of CEA when labelled data for an entirely new class 
is made available. We hide one class during training CEA model (``ABBR'', ``World'', ``Company'' for \textsc{TREC-6}, \textsc{AGNews}, \textsc{DBPedia} respectively). Once training is done, we can just \textit{add} the labelled documents encoded by CEA model into the $k$-nearest neighbour index. It shows that no model updates are required to get quite close in terms of F1-score for each class and overall accuracy compared to a model trained from scratch on the extended corpus.}
\label{table:incremental-learning-new-class}
\vspace{-3mm}
\end{table*}

\noindent{\textbf{Introducing a new class.}}
In this setting, we study a common use-case where the classification need
evolves with time \textit{e.g.,} more classes are introduced at some point into a dataset. 
We simulate this by hiding a class completely during training the CEA model. During inference,
we encode the labelled documents from the new class, and just add their embeddings 
into the $k$-nearest neighbour index like earlier. We present the F1-scores of all individual classes, and overall accuracy on the entire test set in following three settings in Table \ref{table:incremental-learning-new-class} : \textbf{i) Held-out:} The class is completely hidden during training, and inference; \textbf{(ii) Index++:} The class is hidden during training, but added onto the $k$-NN lookup index for inference; \textbf{(iii) Full:} CEA is trained with the full corpus with all classes. Note that, the variants with softmax
classifier do not offer this flexibility.

We observe that CEA indeed shows strong evidence of handling a new
class well without requiring any weight updates. The accuracy it yields after 
extending the index is comparable to a model trained on entire training data 
($0.9620$ vs $0.9720$, $0.9168$ vs $0.9356$, $0.9909$ vs $0.9932$ for 
\textsc{TREC-6, AGNews, DBPedia} respectively). We spot some interesting 
interactions between the classes in this study by observing the ones 
where there is a large change in F1-score after extending the index (denoted
by $\uparrow$ beside). For \textsc{TREC-6}, we find that all the documents 
belonging to ``ABBR'', conflates ``DESC''. We find this consistent with the 
intuitive notion of these two classes. For \textsc{AGNews}, introduction of ``World'' impacts ``Business'', and ``Sci/Tech'' the most. Similarly, in \textsc{DBPedia}, 
the newly introduced class ``Company'' nudges the F1-score of ``Plant'', ``Artist'', ``Building'' by absolute improvements of $0.12, 0.09, 0.10$ respectively. Our further investigation reveals that there is labelling anomaly in the ``Plant'' class (presented in the appendix for the interested readers).\\
\noindent{\textbf{Introducing sub-classes.}}
An existing class might need to be further divided into
multiple sub-classes \cite{wohlwend19acl} with time. The \textsc{TREC} 
corpus lets us study the effect of such scenario. In \textsc{TREC-6} version, all the
questions are annotated with $1$ out of $6$ possible types, whereas in \textsc{TREC-50}, 
the same questions are further divided into $50$ categories. To investigate, how our proposed CEA mechanism would work there, we first train a CEA model on
\textsc{TREC-6}, but during inference we use fine-grained labels from
\textsc{TREC-50} and perform majority voting on them. We observe that CEA yields an 
impressive accuracy of $0.846$ in this task. 

We analyze the embeddings learned by the CEA model to better understand this classification performance in Figure \ref{fig:trec_6_50_transfer}\footnote{enlarged Figure, and full report is provided in appendix.}. We present the $2d$ TSNE plots of the CEA embeddings after training on \textsc{TREC-6} in Figure \ref{subfig:trec_6_tx}. 
On the right (\ref{subfig:trec_50_tx}), we color the same points according to \textsc{TREC-50} classes. Interestingly we observe that, CEA is able to cluster the fine-grained classes together without explicit supervision. This can be explained by the desirable 
property of CEA \textit{i.e.,} it learns the task-specific similarity between texts, 
however still pays attention to the syntactic similarity (Table \ref{tab:label-inconsistency}).

\begin{figure}[t]
    \begin{subfigure}[t]{0.24\textwidth}
        \centering
        \includegraphics[height=1.2in]{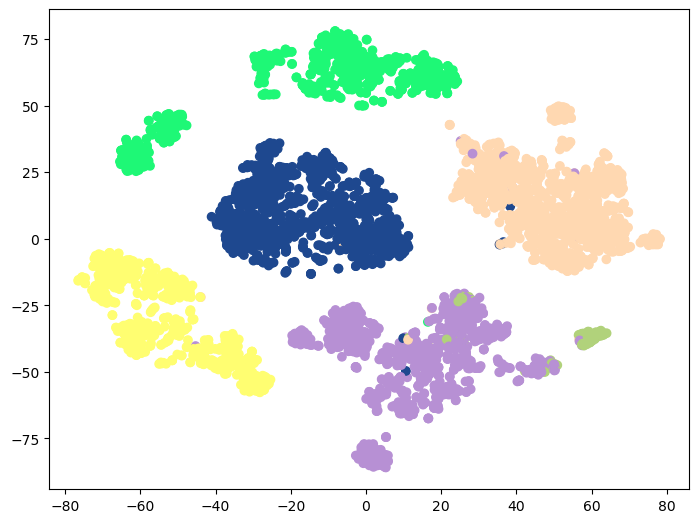}
        \caption{trained on \textsc{TREC-6}}
        \label{subfig:trec_6_tx}
    \end{subfigure}%
    ~
    \begin{subfigure}[t]{0.22\textwidth}
        \centering
        \includegraphics[height=1.2in]{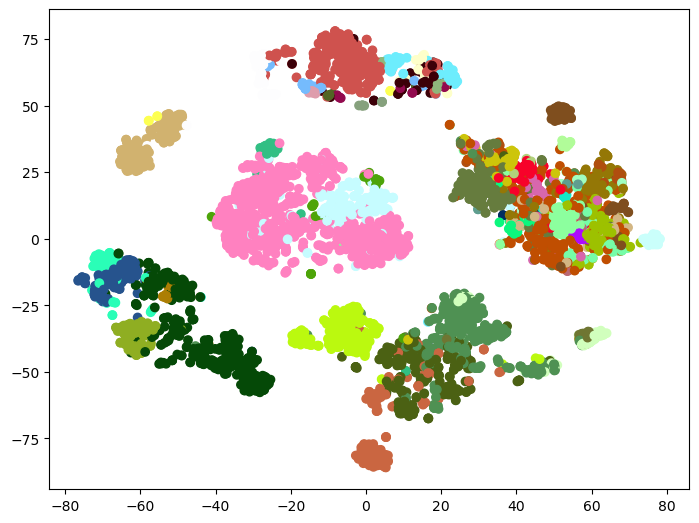}
        \caption{projected on \textsc{TREC-50}}
        \label{subfig:trec_50_tx}
    \end{subfigure}
    \caption{TSNE plot of CEA embeddings for training samples in \textsc{TREC} corpus, coloured according to labels as in a) \textsc{TREC-6}, and b) \textsc{TREC-50}. CEA learns to keep the documents belonging to the sub-classes nearby without explicit supervision.}
    \label{fig:trec_6_50_transfer}
    \vspace{-5mm}
\end{figure}

\noindent{\textbf{Note on alternative objectives:} We have explored other metric learning
approaches for training CEA, such as contrastive loss for pair classification \cite{hadsell2006dimensionality}, and triplet ranking loss \cite{hermans2017defense}. However, our regression based objective yielded the best results, and interpretability of nearest neighbors in our experiments.}
\section{Related Works}
\label{section:related_work}

\textbf{Similarity learning.} 
We build on the work of similarity learning of \citeauthor{reimers2019sentence} which showed that Siamese architectures could be leveraged to capture general semantic similarity across texts. Metric learning based approaches \cite{weinberger09jlmr} were used for few-shot text classification \cite{wohlwend19acl} both in Euclidean and hyperbolic space, and for image class prototype meta-learning \cite{snell2017prototypical}.

\noindent
\textbf{Nearest neighbour approaches.}
Nearest Neighbour has seen applications in NLP in the recent past for sequence labelling \cite{wiseman2019label, sogaard2011semi, chen-chen-2019-k, zhang-etal-2020-discriminative}, as well as 
language modelling \cite{khandelwal20iclr} and question answering \cite{kassner20aclfindings}.
Perhaps the closest ours is work of \citeauthor{wallace18emnlpBBworkshop} where they apply Deep k-Nearest
Neighbors \cite{papernot18arxiv} for text classification. However, they focus on developing reliable uncertainty estimates with simpler
neural models such as CNN \cite{kim-2014-convolutional}, and BiLSTM \cite{sun-et-al-2017}. We believe ours is the first work to benchmark large Transformer based models in conjunction with Nearest Neighbour lookup on a wide range of text classification tasks. 

\noindent
\textbf{Explainability in text classification.}
Majority of work related to explainability (\textit{or} interpretability) of text classification models based on neural networks uses \emph{post-hoc} mechanisms \textit{e.g.,} saliency or attention weights \cite{jain2019attention, wiegreffe2019attention} to attribute part of input to a specific prediction \cite{blackbox18acl}. Our approach is orthogonal to this line of work, as it is based on nearest neighbor classification, which uses \emph{ante-hoc} principles \cite{sokol20fat} by design, such as majority voting for prediction.

\section{Conclusion}
\label{section:conclusion}
In this work, we considered text classification as a task-specific text
similarity problem. To this end, we proposed Class-driven
Embedding Alignment, a training mechanism that brings text documents 
with the same class label closer in embedding space, and pushes others 
further apart. We trained a large Transformer model with our proposed 
approach. We showed the embeddings produced by our model can be used 
in multiple ways to perform accurate text classification, with a softmax 
classifier or non-parametric methods such as $k$-nearest 
neighbor search. Finally, we presented interesting properties of CEA
model, that can be leveraged for introducing transparency in the 
text classification process, as well as imbues flexibility by incrementally 
adding new data without expensive model updates with quantitative 
results and qualitative visualizations of the learned embeddings. 
In the future, we would like to explore how this nearest neighbour 
based approach can be extended for other forms of NLU tasks such as 
textual entailment detection, multi-label text classification where 
the relationship between documents are more complex.

\bibliography{emnlp}
\bibliographystyle{acl_natbib}

\clearpage
\appendix
\section{Appendix}
\label{sec:appendix}

\begin{table}[h]
\centering
\resizebox{0.3\textwidth}{!}{%
\begin{tabular}{ccc}
\hline
Dataset & $k$ & \begin{tabular}[c]{@{}c@{}}Validation\\ Accuracy\end{tabular} \\ \hline
\textsc{TREC-6} & $6$ & $0.9706$ \\
\textsc{TREC-50} & $3$ & $0.9047$ \\
\textsc{AGNews} & $11$ & $0.9767$ \\
\textsc{IMDb} & $21$ & $0.9712$ \\
\textsc{DBPedia} & $13$ & $0.9926$ \\
\textsc{Yelp-full} & $95$ & $0.6832$ \\ \hline
\end{tabular}%
}
\caption{Validation Accuracy for $k$-nearest neighbour search as found during hyper-parameter search within $[1-100]$ with step size of $1$.}
\label{tab:validation-knn}
\end{table}

\begin{table}[h]
\centering
\resizebox{0.45\textwidth}{!}{%
\begin{tabular}{cccc}
\hline
Dataset & CEA & Fine Tuning & NN (inference) \\
\hline
TREC-6 & 32 & 64 & 128 \\
TREC-50 & 32 & 64 & 128 \\
IMDB & 16 & 64 & 128 \\
DBPedia & 16 & 64 & 128 \\
AGNews & 16 & 64 & 128 \\
Yelp-full & 16 & 32 & 128 \\ \hline
\end{tabular}%
}
\caption{Batch-sizes used for different datasets.}
\label{tab:my-table}
\end{table}

\begin{figure}[h]
    \begin{subfigure}[t]{0.5\textwidth}
        \centering
        \includegraphics[height=2.4in]{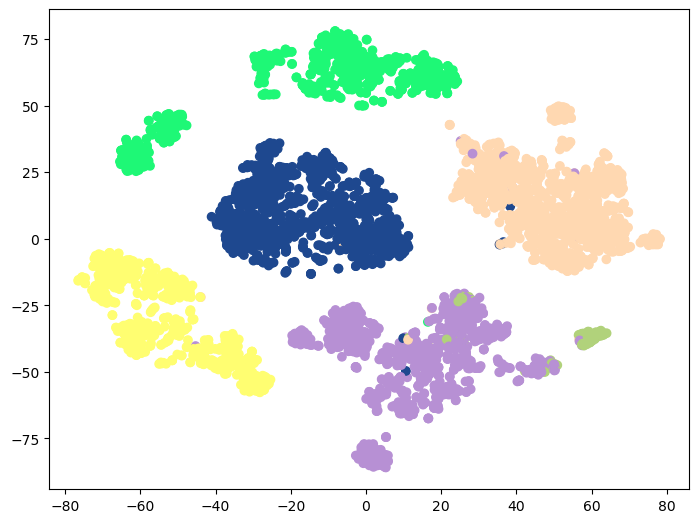}
        \caption{trained on \textsc{TREC-6}}
        \label{subfig:enlarged_trec_6_tx}
    \end{subfigure}%
    
    \begin{subfigure}[t]{0.5\textwidth}
        \centering
        \includegraphics[height=2.4in]{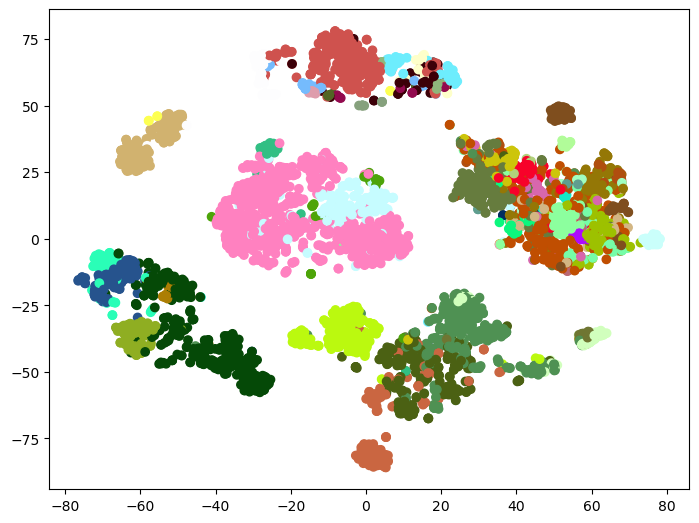}
        \caption{projected on \textsc{TREC-50}}
        \label{subfig:enlarged_trec_50_tx}
    \end{subfigure}
    \caption{(Enlarged Figure \ref{fig:trec_6_50_transfer}) TSNE plot of CEA embeddings for training samples in \textsc{TREC} corpus, coloured according to labels as in a) \textsc{TREC-6}, and b) \textsc{TREC-50}. CEA learns to keep the documents belonging to the sub-classes nearby without explicit supervision, since it is only trained with \textsc{TREC-6}.}
    \label{fig:enlarged_trec_6_50_transfer}
    \vspace{7cm}
\end{figure}

\begin{table*}[]
\centering
\resizebox{0.9\textwidth}{!}{%
\begin{tabular}{|l|l|l|}
\hline
\begin{tabular}[c]{@{}l@{}}Test\\ Document:\end{tabular} & \textbf{``D\&G Bus is a bus operator based in Crewe in Cheshire.''} & \begin{tabular}[c]{@{}l@{}}{\color{red}True label: Company} \\ {\color{blue}Predicted label: Plant}\end{tabular} \\ \hline
Nearest \#1 & \begin{tabular}[c]{@{}l@{}}``VXR is the branding for the high-performance trim specification for models in \\ many of Vauxhall's car range in the United Kingdom.''\end{tabular} & {\color{blue}True label: Plant} \\ \hline
Nearest \#2 & \begin{tabular}[c]{@{}l@{}}``The Paducah \& Louisville Railway (reporting mark PAL) is a Class II railroad \\ that operates freight service between Paducah and Louisville Kentucky. {[}..{]}''\end{tabular} & {\color{blue}True label: Plant} \\ \hline
Nearest \#3 & \begin{tabular}[c]{@{}l@{}}``Roush Fenway Racing (originally Roush Racing) is a racing team competing in \\ NASCAR racing. As one of NASCAR's largest premier racing teams Roush {[}..{]}''\end{tabular} & {\color{blue}True label: Plant} \\ \hline
Nearest \#4 & \begin{tabular}[c]{@{}l@{}}``The Belgrano Norte line is a commuter rail service in Buenos Aires Argentina run \\ by the private company Ferrovías since 1 April 1994. This service previously had {[}..{]}''\end{tabular} & {\color{blue}True label: Plant} \\ \hline
\end{tabular}%
}
\caption{Nearest Neighbours for a test document in \textsc{DBPedia} dataset when a new class ``Company'' is added to the index. Note that, all the nearest neighbours indeed are similar. However, they carry the wrong ground-truth labels, affecting the final prediction according to majority voting.}
\label{tab:dbpedia-anomalies}
\end{table*}

\begin{figure*}[]
    \centering
    \includegraphics[height=8in]{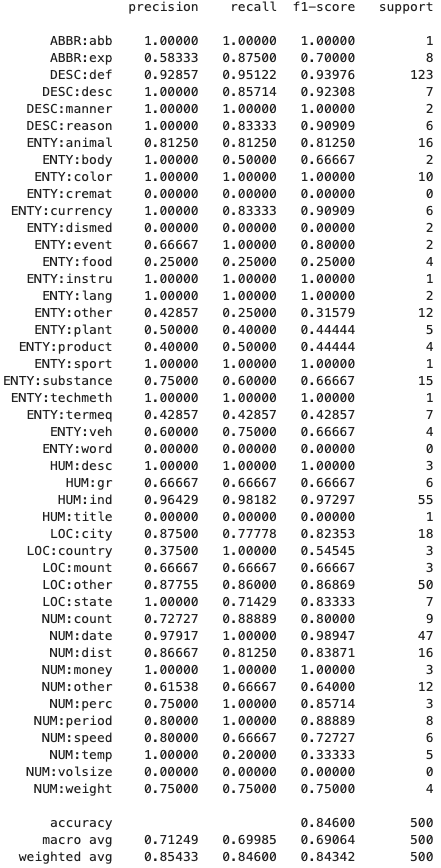}
    \caption{Full classification report for all $50$ classes for the \textsc{TREC-6} to \textsc{TREC-50}
    transfer experiment.}
    \label{fig:trec6_50_report}
\end{figure*}


\end{document}